# A Hybrid Distribution Feeder Long-Term Load Forecasting Method Based on Sequence Prediction

Ming Dong, *Senior Member, IEEE* and L.S.Grumbach

***Abstract*—Distribution feeder long-term load forecast (LTLF) is a critical task many electric utility companies perform on an annual basis. The goal of this task is to forecast the annual load of distribution feeders. The previous top-down and bottom-up LTLF methods are unable to incorporate different levels of information. This paper proposes a hybrid modeling method using sequence prediction for this classic and important task. The proposed method can seamlessly integrate top-down, bottom-up and sequential information hidden in multi-year data. Two advanced sequence prediction models Long Short-Term Memory (LSTM) and Gated Recurrent Unit (GRU) networks are investigated in this paper. They successfully solve the vanishing and exploding gradient problems a standard recurrent neural network has. This paper firstly explains the theories of LSTM and GRU networks and then discusses the steps of feature selection, feature engineering and model implementation in detail. In the end, a real-world application example for a large urban grid in West Canada is provided. LSTM and GRU networks under different sequential configurations and traditional models including bottom-up, ARIMA and feed-forward neural network are all implemented and compared in detail. The proposed method demonstrates superior performance and great practicality.**

## I. Introduction

Different from short-term load forecast (STLF), long-term load forecast (LTLF) problem refers to forecasting electrical power demand in more than one-year planning horizon for different parts of a power system [1-3]. It is the essential foundation of system planning activities in utility companies. LTLF establishes a necessary understanding of system adequacy for reliably supplying power to meet future customer demand. Peak demand is often used as the forecast target because it represents the worst case scenario and needs to be tested against system capacity constraints.

Long-term forecast of peak demand at distribution feeder level is especially important because it is used as the input to assess the power delivery capacity during normal operation and the restoration capability during system contingencies for the next few years. Only after proper forecast and assessment, utility companies can reasonably plan long-term infrastructure upgrades and modifications [1-3]. Examples are transferring loads between feeders, adding feeder tie-points, building new feeders, installing new transformers, building new substations and etc. Therefore, distribution feeder LTLF significantly affects the reliability of future grid, the satisfaction of utility customers, the capital investment and financial outcome of utility companies.

In general, LTLF methods can be grouped into the following three categories [4-6]:

*1) Top-down Forecasting:* this category focuses on forecasting electricity usage at a group-level such as the load of all customers or the load of residential sector in a region [4]. Some methods use single or combinations of univariate regression models such as ARIMA to analyze the trend of loading change [7-9]. These methods only analyze the temporal loading variable itself and are generally unacceptable for LTLF because long-term load change is strongly driven by external variables such as economy, population and weather. To overcome this problem, some methods use multivariate regression models such as feed-forward neural network (FNN) to analyze those external variables and their relationships with the loading change [10-14]. The advantage of these methods is the statistical explicability. Utility companies can now forecast and explain future load change based on other variables forecasted by government or third-party agencies. These methods work well for regional or group-level load forecast but can be challenging when applied to system components such as individual distribution feeders. This is because the top-down process of allocating group-level load to individual members is subjective. There is no clear way to reconcile the group-level information with member-level information. It is also unrealistic to assume all members simply comply with the group-level load behavior. In reality, a distribution feeder's peak demand can be greatly affected by its large loads and significantly deviates from its regional load behavior. Therefore, in practice top-down forecasting only provides an overall reference for manual check and adjustment of member-level forecast [4,6].

*2) Bottom-up Forecasting:* in contrast to top-down forecasting, this category requires gathering bottom customer load information to build a higher level forecast. One approach of information gathering is conducting utility surveys or interviews. Long-term load information such as

M. Dong is with Department of System Planning, ENMAX Power Corporation, Calgary, AB, Canada, T2G 4S7 (e-mail: mingdong@ieee.org)

L.S.Grumbach is with Auroki Analytics, Vancouver, BC, Canada, V7L 1E6 (e-mail: lsgrumbach@auroki.com)

expected sizes of new loads, load maturation plan and/or long-term production plan is obtained, summarized and estimated as annual loading change. In practice, this is only done for large customers since those customers can substantially affect the feeder-level loading and it is too costly to gather load plans from all customers [4]. Despite the tremendous effort required to communicate with major residential developers, commercial and industrial customers, inaccurate forecast often occurs with this approach due to unreliable customer information and change of customer plans over the forecasting horizon. As an alternative to surveys or interviews, some methods rely on the use of sub-load profiles [15-16]. Sub-load profiles are forecasted individually or by clusters and then aggregated to a higher level. This is an effective approach for SLTF. However, missing statistical analysis of load variation driven by external factors made it unreliable for long-term forecast tasks.

*3) Hybrid Forecasting:* this approach attempts to combine the advantages of top-down and bottom-up forecasting. Unfortunately not many research works were found in this direction. One example is the statistically-adjusted end-use model for household-level load forecast [17]. It combines top-down weather, household and economic information with bottom-up appliance information to forecast household-level load. No literature was found for distribution feeder LTLF using similar methods.

In response to the above literature findings, the first contribution of this paper is the establishment of a hybrid forecasting method that can effectively combine regional economic, demographic and temperature information with feeder-level load information in one mathematical model. As a result, this model can reflect the effects of overall regional drivers on feeder peak demand; it can also reflect large customer load change, load composition, Distributed Energy Resources (DER) and Electric Vehicle (EV) adoption information specific to individual feeders.

The second contribution of this paper is the adoption of sequence prediction models to extract and utilize the long-term sequential patterns of peak demand to improve forecast accuracy. Sequence prediction is the problem of using historical sequence information to predict the next value or values in a sequence [18]. Both LSTM and GRU networks are commonly used advanced sequence prediction models. Compared to ARIMA, they support input and output with multiple features; compared to a standard recurrent neural network (RNN), they solved the vanishing and exploding gradient problems and are therefore much more stable [19-24]. In a way, these models can combine the advantages of univariate trending analysis and complex multivariate regression models. In recent years, researchers applied them to classic time-series problems such as stock, weather forecasting and machine translation [25-27]. They often outperform traditional regression models such as FNN in these tasks. It was not until very recently that some researchers started to apply LSTM and GRU to STLF problems in power systems [28-30]. The application of LSTM and GRU to LTLF problems has not been found through literature review. This paper aims to fill this research gap and explore the use of LSTM and GRU networks under different sequential configurations for one of the most classic and important long-term forecasting tasks – forecasting individual feeder long-term peak demand.

The structure of the proposed modeling method is shown in Fig.1. Raw top-down features (related to economy, population and temperature), raw bottom-up features (related to customer load and DER/EV adoption), and previous-year peak demand are all fed into a feature engineering module. For feature engineering, the concept of virtual feeder is proposed to eliminate the data corruption resulted from historical load transfer events between feeders; feature normalization is applied to normalize different types of features to the same numerical scale; then principal component analysis is applied to reduce the dimensionality of highly correlated features to improve model training efficiency and avoid over-fitting problems. After the step of feature engineering, the dataset is constructed to a unique multi-time step format under either many-to-many or many-to-one sequential configurations. The dataset is also split into training set and test set for training and evaluation purposes. After model evaluation and network parameter tuning, a reliable sequence prediction model for distribution feeder LTLF is established and can be used for future forecast.

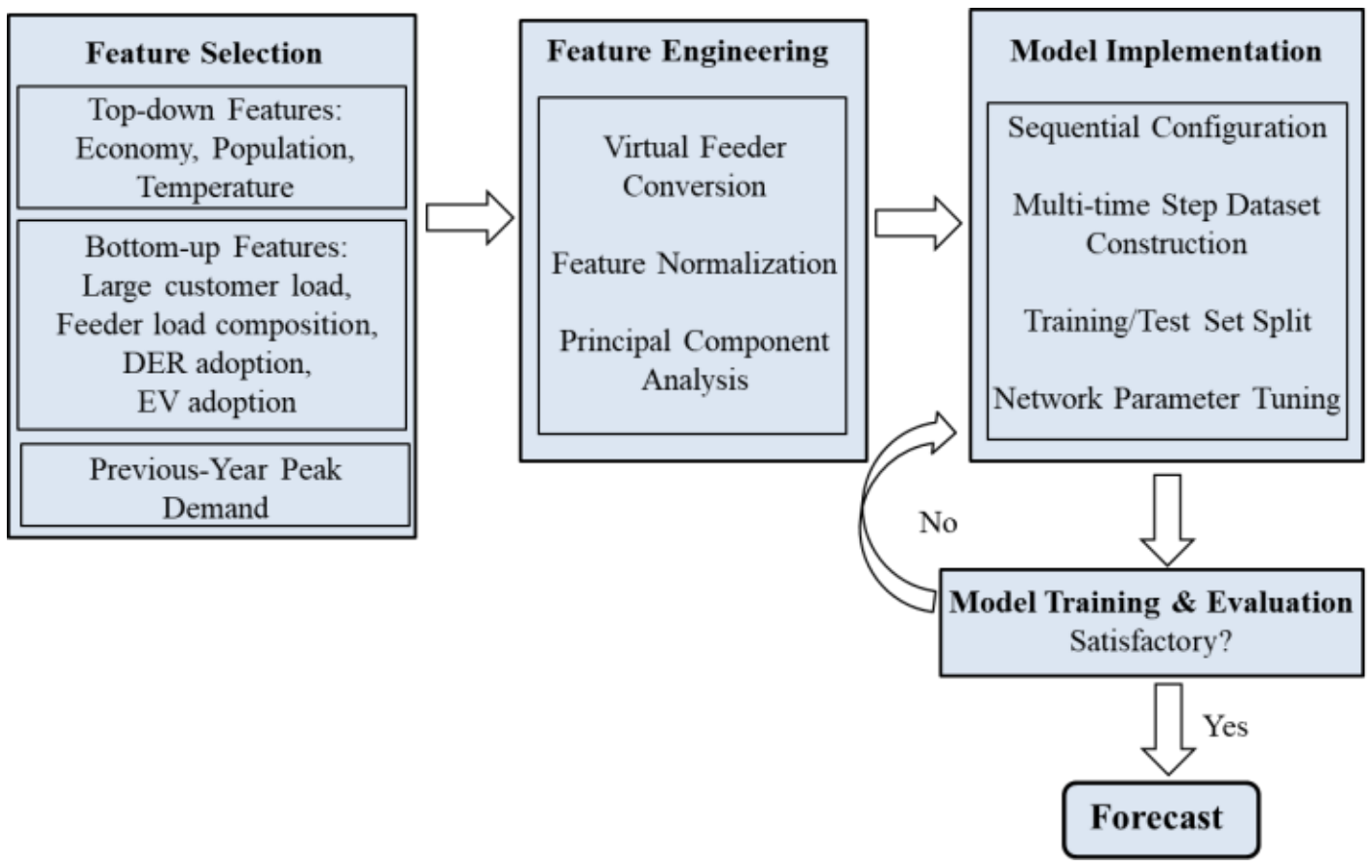

Fig.1. Workflow of the proposed modeling method

This paper firstly introduces the theories of LSTM and GRU networks. It then elaborates the workflow of feature selection, feature engineering and model implementation as shown in Fig.1. In the end, a real-world application to a large urban grid in West Canada with 289 feeders is presented and discussed in detail. As part of the model evaluation, the proposed method is compared to traditional methods including bottom-up, ARIMA and FNN. It demonstrates superior performance over all of them.

## II. Introduction of LSTM and GRU Models

This section provides a brief introduction to LSTM and GRU models and establishes the mathematical foundation for the proposed method. Since LSTM and GRU models are both based on RNN, this section firstly reviews standard RNN and then explains the working principles of LSTM and GRU and their advantages over standard RNN.

### A. Recurrent Neural Network

As shown in Fig.2, a RNN is a group of FNNs where hidden neurons of the FNN at a previous time step are connected with the hidden neurons of the FNN at the following time step. The state of hidden neurons $h_t$ is generated from $h_{t-1}$ at the previous time step and the current data input $X_t$ by applying weights $W_R$ and $W_{in}$. At each time step $t$, an output $Y_t$ is produced. This process continues for the next time step and so on. This way, RNN is able to make use of sequential information and does not treat one time step as an isolated point. This nature made RNN suitable for forecasting tasks where the output of current time step is not only based on the current input but also the information from previous time steps. Taking LTLF problem as an example, the current power demand is often not only related to the current year but also related to the conditions and momentum of the past few years.

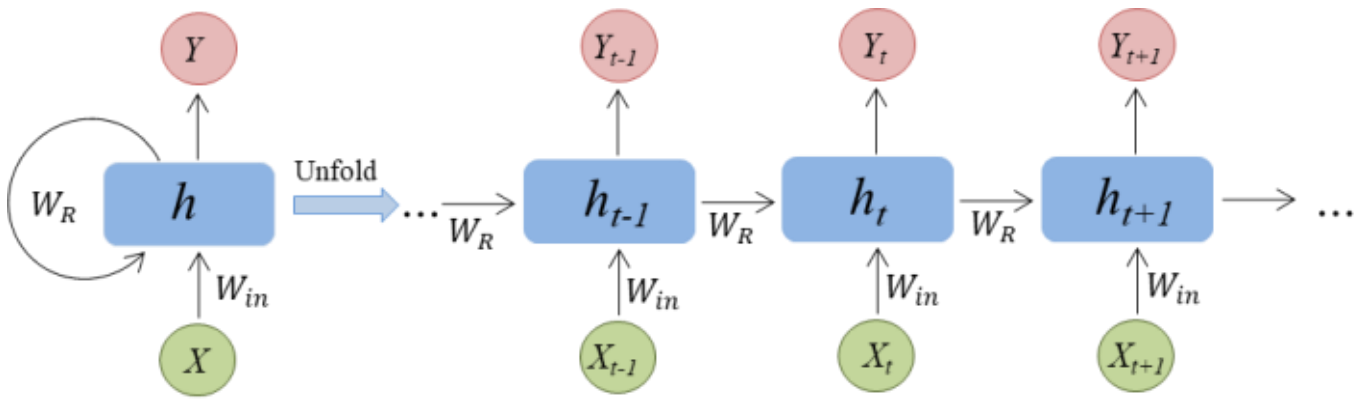

Fig.2. Illustration of an unfolded RNN

Although RNN has a better performance than FNN when dealing with time-series data, the training of a RNN can be unstable due to an intrinsic problem called vanishing/exploding gradient. This problem is caused by the long distance during backpropagation of network loss from one FNN to another FNN a few time steps ago [19-20]. During backpropagation of RNN, gradient value may become very small and the training process loses traction; gradient value can also become very large and lead to overly large change of weights between updates.

### B. LSTM Model

To solve the vanishing/exploding gradient problem, LSTM model was proposed to improve the RNN structure [20-21]. Compared to standard RNN, LSTM introduces a specially designed LSTM unit to sophisticatedly control the flow of hidden state information from one time step to the next. The structure of LSTM unit is shown in Fig.3.

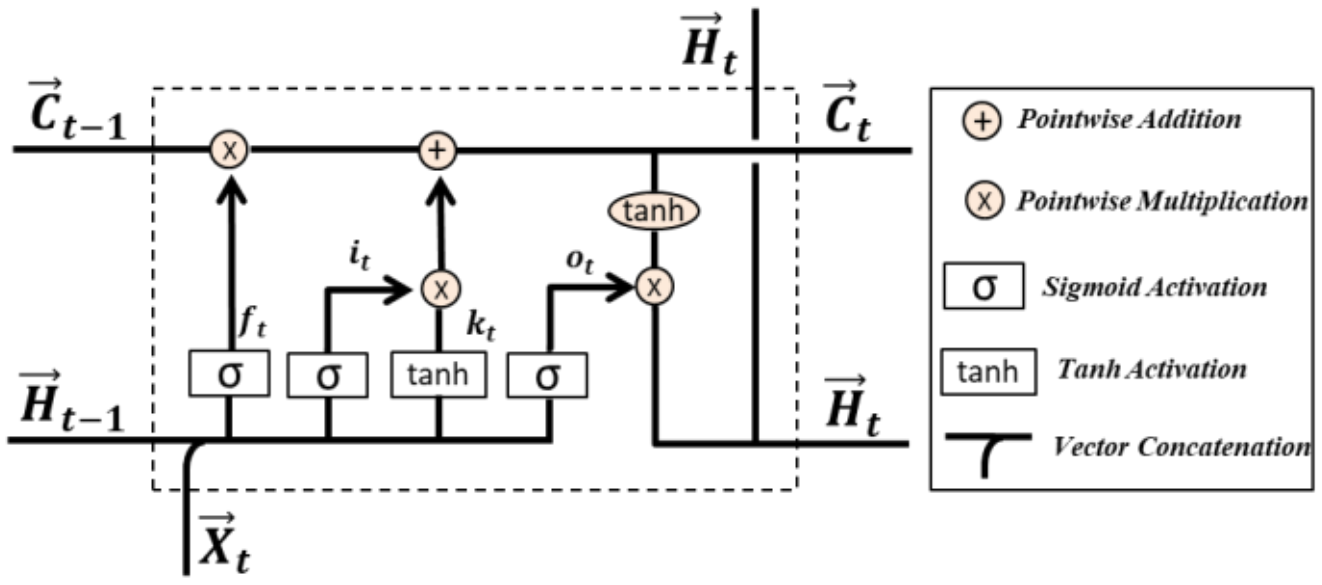

Fig.3. A LSTM unit diagram

In Fig.3, $\vec{X}_t$ and $\vec{H}_t$ are the input vector and network hidden state vector at time step $t$. $\vec{C}_t$ is a vector stored in an external memory cell. This memory cell carries information between time steps, interacts with input vector and hidden state vector and gets updated from one time step to the next. This interaction is completed through three control gates: forget gate, input gate and output gate.

A forget gate element $f_t$ is calculated by:

$$f_t = \sigma\left(\vec{W}_f \cdot \left[\vec{H}_{t-1,}\vec{X}_t\right] + b_f\right) \quad (1)$$

where $\left[\vec{H}_{t-1,}\vec{X}_t\right]$ is the concatenated vector of previous hidden state vector $\vec{H}_{t-1}$ and the current input vector $\vec{X}_t$; $\sigma$ is the sigmoid activation function; $\vec{W}_f$ and $b_f$ are the weight vector and bias. They are determined through network training. The sigmoid activation function outputs a value $f_t$ between 0 and 1. In the forget gate vector, each element $f_t$ controls how the corresponding element in the cell state vector $\vec{C}_t$ gets kept or forgotten. 1 means keeping the element unchanged and 0 means zeroing out the element. This is achieved by pointwise multiplying forget gate vector $\vec{f}_t$ by $\vec{C}_t$ and is mathematically given later in (4).

Following the information flow in Fig.3, a temporary cell state element $k_t$ is calculated by:

$$k_t = tanh(\vec{W}_k \cdot \left[\vec{H}_{t-1,}\vec{X}_t\right] + b_k) \quad (2)$$

where $\left[\vec{H}_{t-1,}\vec{X}_t\right]$ is the concatenated vector of previous hidden state vector $\vec{H}_{t-1}$ and the current input vector $\vec{X}_t$; *tanh* is the tanh activation function and outputs a value between -1 and 1; $\vec{W}_k$ and $b_k$ are the weight vector and bias.

In parallel with calculating $K_t$, the input gate $i_t$ is calculated by:

$$i_t = \sigma(\vec{W}_i \cdot \left[\vec{H}_{t-1,}\vec{X}_t\right] + b_i) \quad (3)$$

where $W_i$ and $b_i$ are the weight vector and bias of $i_t$.

Eventually the new cell state $\vec{C}_t$ at time step $t$ is updated with previous cell state vector $\vec{C}_{t-1}$, forget gate vector $\vec{f}_t$, input gate vector $\vec{\iota}_t$ and temporary cell state vector $\vec{k}_t$ by using pointwise multiplication and addition:

$$\vec{C}_t = \vec{f}_t \otimes \vec{C}_{t-1} + \vec{\iota}_t \otimes \vec{k}_t \quad (4)$$

This new cell state further determines the hidden state in the current neural network at time step $t$ through the output gate $o_t$. Similar to $f_t$ and $i_t$, $o_t$ is calculated by:

$$o_t = \sigma\left(\vec{W}_o \cdot \left[\vec{H}_{t-1,}\vec{X}_t\right] + b_o\right) \quad (5)$$

Then, hidden state $H_t$ at the current time step $t$ is calculated by pointwise multiplying output gate vector $\vec{o}_t$ by the *tanh* function of $\vec{C}_t$:

$$\vec{H}_t = \vec{o}_t \otimes tanh(\vec{C}_t) \quad (6)$$

Through (1) to (6), the current hidden state $\vec{H}_t$ is calculated with the use of $\vec{C}_{t-1}$ and $\vec{H}_{t-1}$ from the previous time step as well as the current input $\vec{X}_t$. $\vec{H}_t$ is then used to produce network output $Y_t$ at the current time step.

LSTM model inherits the advantages of RNN in dealing with temporal forecast problems and also solves the vanishing/exploding gradient problem by using the LSTM unit.

### C. GRU Model

GRU model is a newer sequence prediction model invented in 2014 by Cho et al. when they researched machine translation problems [22]. Compared to LSTM, GRU eliminates the use of the memory cell and uses only hidden state to carry information flow. It also merges the forget and input gates into a single update gate. Generally, GRU is more efficient than LSTM due to fewer gates being used in the process. However, from the accuracy perspective, one model is not always better than the other [23], except for certain language modeling tasks [24]. As a result, in practice LSTM and GRU models can be selected using a trial and error approach for a specific problem or a specific dataset. This is also the approach suggested later in this paper. The structure of a GRU unit is shown in Fig.4.

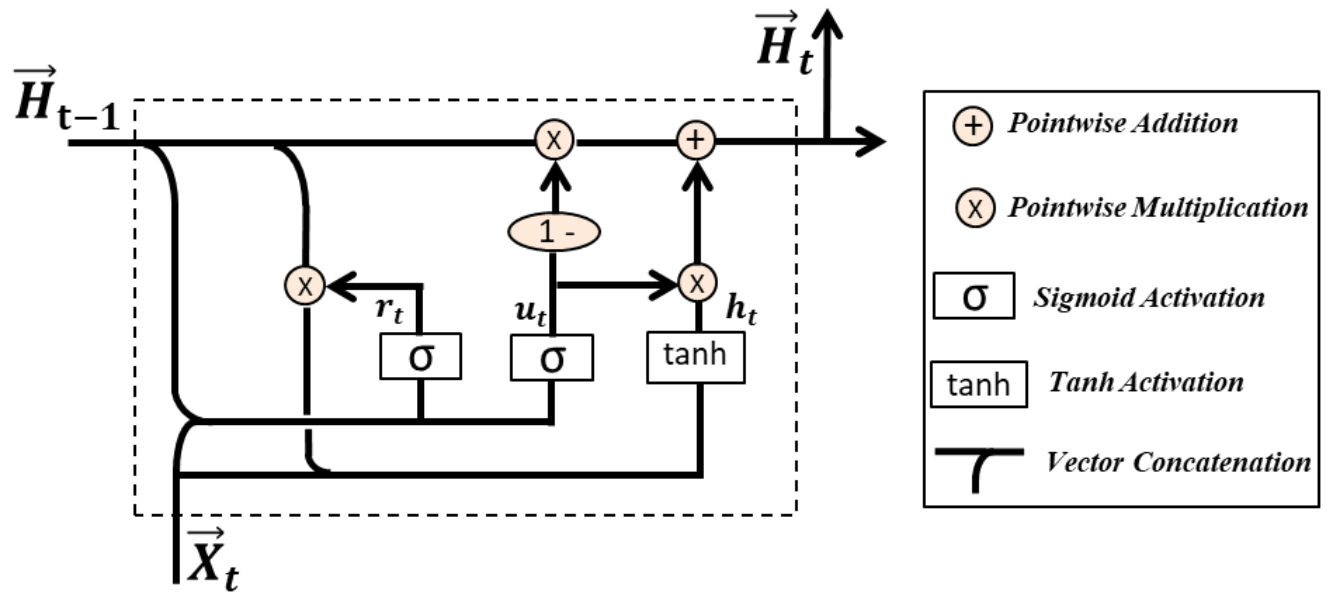

Fig.4. A GRU unit diagram

A reset gate element $r_t$ is calculated by:

$$r_t = \sigma(\vec{W}_r \cdot [\vec{H}_{t-1,}\vec{X}_t] + b_r) \quad (7)$$

where $[\vec{H}_{t-1,}\vec{X}_t]$ is the concatenated vector of previous hidden state vector $\vec{H}_{t-1,}$ and the current input vector $\vec{X}_t$; $\vec{W}_r$ and $b_r$ are the weight vector and bias.

An update gate $u_t$ element is calculated by using the same input and activation function from (7):

$$u_t = \sigma(\vec{W}_u \cdot [\vec{H}_{t-1,}\vec{X}_t] + b_u) \quad (8)$$

with different weight vector $W_u$ and bias $b_u$.

Then following the information flow illustrated in Fig.4, a temporary value $h_t$ is calculated by:

$$h_t = tanh(\vec{W}_h \cdot [\vec{r}_t \otimes \vec{H}_{t-1,}\vec{X}_t] + b_h) \quad (9)$$

Finally, the hidden state vector $\vec{H}_t$ at the current time step *t* is calculated with previous hidden state vector $\vec{H}_{t-1}$, update gate vector $\vec{u}_t$ and temporary vector $\vec{h}_t$ by using pointwise multiplication and addition :

$$\vec{H}_t = (1 - \vec{u}_t) \otimes \vec{H}_{t-1} + \vec{u}_t \otimes \vec{h}_t \quad (10)$$

Through (7)-(10), hidden state $\vec{H}_t$ is updated from one time step to the next. It affects the neural network output at each time step.

Overall, LSTM and GRU models have more complicated structures and more internal parameters than standard RNN and FNN. As a result they will need longer training time. However, they both solved the vanishing/exploding gradient problem and are reliable sequence prediction models.

## III. Feature Selection

Feature selection is normally the first step of building a machine learning model [31]. By employing domain knowledge, useful raw features related to the problem are analyzed and selected. In the proposed hybrid model, both top-down features and bottom-up features related to distribution feeder peak demand are selected. They are elaborated as below.

### *A. Top-down Features*

Top-down features describe the overall drivers in the forecasted region. Annual economic, population and temperature features are considered in the model. The historical and future economic and population features can often be obtained from third-party consultants or government agencies [32]. The historical temperatures can be obtained from weather statistics datasources [33]. Future long-term temperatures, however, are difficult to forecast. In practice, depending on the conservativeness of system planning, they can be normalized to either the average or extreme temperature point observed in a region from the past few years. This is further explained in Section V-E.

*1) Economic Features:* Different from short-term power demand, long-term power demand is largely driven by local economy. Annual real GDP growth (%) is the nominal GDP that excludes the effect of inflation rate; total employment growth (%) is another important economic feature. Higher employment means more people hired in the commercial and industrial sectors and may potentially use more electricity; housing starts is the number of residential units that are started to construct in a year in a region. This indicator is related to the increase of residential electricity usage and can be selected when available.

Additional supplementary economic features include industrial production indexes and commodity prices [32]. They are more related to industrial loads and can be selected according to the industry composition in the forecasted region.

*2) Population Features:* Population size significantly affects the residential load growth. Even when the economy slows down, a stable population size can still support stable residential loading level. This is because most of the residential electricity demand comes from everyday household activities such as lighting, cooking, laundry and so on. These activities are relatively immune to economic condition. Furthermore, population growth can result in the residential development which requires electricity supply during construction and after possession. In addition, as part of the population, labor force in return affects economic activities and is related to the total employment growth. Therefore, population growth (%) is selected in this work. Another useful population feature for some regions is net migration [31]. It is the annual difference between the number of immigrants and emigrants. This feature excludes the population change due to natural birth and death and is more closely related to regional economic attractions. It can be considered for regions with frequent population migration.

*3) Max/Min Temperature:* Depending on forecasting summer peak demand or winter peak demand, the highest temperature during summer or the lowest temperature during winter is selected for each year. This is because summer peak demand and winter peak demand often align with temperature extremes due to cooling and heating electricity use [34-35]. This correlation can be especially significant for summer because cooling almost always relies on electricity usage whereas heating may rely on other energy sources such as natural gas. Both the peak temperature value and the temperature change from the previous year are selected.

### *B. Bottom-up Features*

Bottom-up features describe the detailed feeder-level load information. Large customer net load change, feeder load composition and DER/EV adoption growth are considered in the proposed models.

*1) Large Customer Net Load Change*: this feature is the estimated net load change of all large customers on the feeder. Examples of large customers can be factories, shopping malls, office buildings and new residential developments. For a future year, the load information from each large customer can be collected through utility survey or interview. Some may report growth while some may report reduction. The aggregated net change is the summation of all these reported load changes from large customers on a feeder. Sometimes utility companies may decide to further adjust the reported load changes based on their own understanding in case customers report unrealistic information.

*2) Feeder Load Composition (%):* Distribution feeders have different types of loads on them and they respond to top-down features in different ways. For example, residential feeders are more related to temperature and population while industrial loads are more related to economy. Feeder load composition features can reflect this difference. Residential peak load percentage of a feeder is calculated by:

$$R = \frac{\sum_{i=1}^{n} L_i^R}{P_F} \times 100\% \qquad (11)$$

where $P_F$ is the peak loading of the feeder in the summer or winter of previous year; $L_i^R$ is the loading of residential load $i$ at the feeder's peaking time for $P_F$; $n$ is the total number of residential loads on this feeder.

Similarly, commercial peak load percentage of a feeder is calculated by:

$$C = \frac{\sum_{i=1}^{m} L_i^C}{P_F} \times 100\% \qquad (12)$$

where $L_i^C$ is the loading of commercial load $i$ at the feeder's peaking time for $P_F$; $m$ is the total number of commercial loads on this feeder.

The industrial load percentage can be calculated in a similar way. It can also be calculated by:

$$I = 1 - R - C \qquad (13)$$

In actual application, only two features out of three need to be selected because they are mathematically correlated with the third feature as (13) shows.

*3) DER Adoption Growth:* Customer adoption of DER may reduce the peak demand of feeders. Two residential feeders with similar numbers of customers may have significantly different peak demand when they have very different DER adoption rates. In regions where DER is a concern, features such as the forecasted number of new DER installations or DER MW output can be selected. DER adoption growth itself can be forecasted based on customer propensity analysis using methods such as [36] and is not discussed in this paper.

*4) EV Adoption Growth:* Customer adoption of EV may increase the peak demand due to battery charging activities. In regions where EV is a concern, features such as the forecasted number of newly purchased EVs can be selected. EV adoption growth can be forecasted based on customer propensity analysis using methods such as [37] and is not discussed in this paper.

### *C. Previous-Year Peak Demand*

Depending on forecasting summer peak or winter peak, the previous year's summer or winter peak demand is required in this model. The Previous-Year Peak Demand feature serves as a baseline while all the discussed top-down and bottom-up features except feeder load composition (%) focus on the change of the following year. Together, all these features lead to the forecast of the following peak demand.

The features discussed in this section are summarized in Table I. All the mandatory features are important because these are the primary features representing different factors that affect feeder loading. They are also readily available from a utility application perspective. Optional features are specific to regions and may be included if applicable. Mathematically, whether to add an optional feature can also be determined by comparing the forecast accuracy before and after adding it to the model.

TABLE I: FEATURES CONSIDERED IN THE PROPOSED METHOD

| Feature Name | Category | Requirement |
|---|---|---|
| Real GDP Growth (%) | Top-down | Mandatory |
| Total Employment Growth (%) | Top-down | Mandatory |
| Population Growth (%) | Top-down | Mandatory |
| Max/Min Temperature | Top-down | Mandatory |
| Max/Min Temperature Change | Top-down | Mandatory |
| Large Customer Net Load Change | Bottom-up | Mandatory |
| Residential Peak Load Percentage | Bottom-up | Mandatory |
| Commercial Peak Load Percentage | Bottom-up | Mandatory |
| Previous-Year Peak Demand | Baseline | Mandatory |
| # of Housing Starts | Top-down | Optional |
| Industrial Production Index | Top-down | Optional |
| Commodity Price | Top-down | Optional |
| Net Migration | Top-down | Optional |
| DER Adoption Growth | Bottom-up | Optional |
| EV Adoption Growth | Bottom-up | Optional |

## IV. FEATURE ENGINEERING

Feature engineering is the step to transform raw features discussed in Section III to proper features that can be fed into the proposed models for training [31]. The purpose of feature engineering is to eliminate data noise, reduce model complexity and improve model accuracy.

### *A. Virtual Feeder Features*

In practice, one significant type of data noise that affects feeder peak demand over a long term comes from the load transfer events between adjacent feeders. A certain amount of customers can be switched between adjacent feeders. This is often driven by system operational needs. For example, when feeder A's loading is getting close to its capacity constraint, it is decided to transfer the customers located on a feeder branch of feeder A to its adjacent feeder B so that both feeder A and B can continue to reliably supply their customers. In this case, load transfer creates a sudden load drop on feeder A and a sudden load rise on feeder B. This change deviates from the previous loading trend on both feeders and has nothing to do with any top-down and bottom-up features discussed in Section III. Another example is maintenance driven load transfer. Feeder A may need to be de-energized to maintain, replace or upgrade its substation breaker, conductors and cables. During this type of maintenance work, feeder A needs to be sectionalized and customers in each feeder section are transferred to multiple adjacent feeders. Load transfer can be done through switching pre-installed branch switches and feeder-tie switches as illustrated in Fig.5.

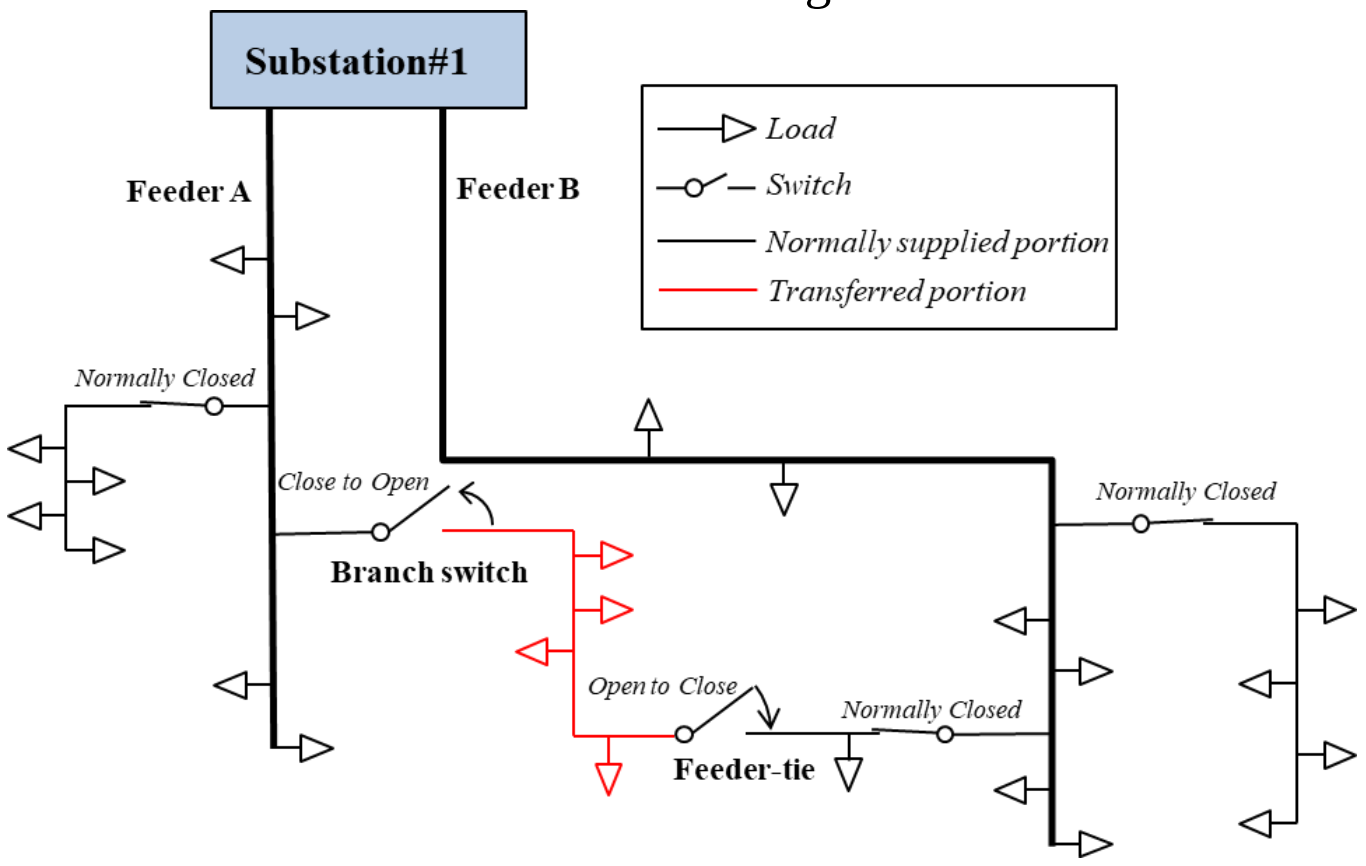

Fig.5. An example of load transfer from feeder A to feeder B

Load transfer is an almost inevitable event in distribution grid. Over a long period of time such as a few years, a significant portion of distribution feeders can be affected by this event. Load transfer events create data noise and significantly reduce the accuracy of the model if raw features are directly used for modeling.

To overcome this problem, this paper proposes a concept called virtual feeders. This unique technique will ensure the continuity of feeder loading trend in the dataset. For one area, a virtual feeder can be created and it is the average of the adjacent feeders which had load transfer events in the model training period. Instead of using features of individual feeders in this area as training records, the features of virtual feeder are generated and used. The Previous-Year Peak Demand feature of the virtual feeder can be estimated by:

$$P_V = \frac{\sum_{i=1}^{p} P_i}{p}, p \geq 2 \tag{14}$$

where $P_i$ is the Previous-Year Peak Demand feature of adjacent feeder $i$ involved in historical load transfers during the model training period; $P_V$ is the Previous-Year Peak Demand feature of the virtual feeder; $p$ is the number of adjacent feeders that have switching events during the model training period. $p$ is normally 2 but can be greater than 2 for multi-feeder switching during feeder maintenance activities.

Similarly, large customer net load change feature $LC$ of virtual feeder can be calculated by:

$$LC = \frac{\sum_{i=1}^{p} LC_i}{p} \tag{15}$$

DER and EV adoption growth on the virtual feeder can be calculated by:

$$D = \frac{\sum_{i=1}^{p} D_i}{p} \tag{16}$$

$$E = \frac{\sum_{i=1}^{p} E_i}{p} \tag{17}$$

where $D_i$ and $E_i$ are the DER and EV growth features of feeder *i*.

Residential Peak Load Percentage and Commercial Peak Load Percentage $R$ and $C$ of the virtual feeder can be estimated by:

$$R_V = \frac{\sum_{i=1}^{p} R_i P_i}{p P_V} \times 100\% \tag{18}$$

$$C_V = \frac{\sum_{i=1}^{p} C_i P_i}{p P_V} \times 100\% \tag{19}$$

where $R_i$ and $C_i$ are the residential and commercial peak load percentage features of feeder *i*.

The top-down features in Table I do not need to be updated for virtual feeders as they represent the governing regional characteristics. By creating virtual feeder features, the data noise coming from load transfer events can be effectively eliminated.

### B. Feature Normalization

This is a necessary step because the features discussed in Section III use different units and have large magnitude differences among them. There are many ways of normalizing raw features [31], for example, the Min-Max normalization can normalize features to the value range of [0,1]. It is given by:

$$X_{norm} = \frac{X_{Raw} - MIN}{MAX - MIN} \tag{20}$$

where for a specific feature, $MAX$ is the maximum observed value in this feature; $MIN$ is the minimum observed value in this feature.

### C. Principal Component Analysis

Table I contains many economic and population features. These features emphasize different aspects but are highly correlated. For example, Net Migration can incent Real GDP Growth and lead to Housing Starts Growth; Total Employment Growth often goes hand-in-hand with Real GDP Growth. These features are not independent and can be aggregated using principal component analysis (PCA) [31]. This is important because LTLF has to rely on annual data points. Not like STLF which often uses hourly data points, annual data points are limited in number. Reducing feature dimensionality can improve model accuracy and avoid over-fitting problems.

PCA is defined as an orthogonal linear transformation that maps the data with multiple variables to a new coordinate system. In this new coordinate system, coordinates are orthogonal (independent) to each other and the greatest data variance direction aligns with the first coordinate (i.e. the first principal component), the second greatest variance direction aligns with the second coordinate, and so on. Mathematically, the transformation is defined as:

$$\begin{cases} \mathbf{T} = \mathbf{XP} \\ \mathbf{X^T X} = \mathbf{PDP^{-1}} \end{cases} \tag{21}$$

where $\mathbf{X}$ is the normalized mean-shifted data matrix with $k$ feature columns (each column is subtracted by the column mean); $\mathbf{P}$ is a $k$-by-$k$ matrix whose columns are eigenvectors of $\mathbf{X^T X}$; $\mathbf{D}$ is the diagonal matrix with eigenvalues $\lambda_j$ on the diagonal and zeros everywhere else; **T** is the new data matrix in which each column is an independent feature projected to a principal component.

After the transformation, $t$ independent features can be selected from **T** based on the Proportion of Variance Explained (PVE). PVE of $t$ features indicates the amount of information (variance) attributed to these features and is mathematically given as below:

$$PVE = \sum_{i=1}^{t} \lambda_i / \sum_{j=1}^{k} \lambda_j \tag{22}$$

An example of transforming four economic-population features to two independent features *EP1* and *EP2* along two orthogonal principal components is shown in Table II. Two features are selected because their corresponding PVE value using (22) is 97.1%. This means most of the information from the original four features can be kept in the two newly constructed features *EP1* and *EP2*.

TABLE II: AN EXAMPLE OF PRINCIPAL COMPONENT ANALYSIS

| Real GDP Growth (%) | Total Employment Growth (%) | Population Growth (%) | Net Migration ('000 Persons) | New Features After PCA | |
|---|---|---|---|---|---|
| | | | | EP1 | EP2 |
| 14.2 | 4.9 | 2.9 | 17.6 | -0.64 | 0.44 |
| 9.1 | 2.7 | 2.2 | 12.4 | -0.16 | 0.31 |
| -2.5 | -0.5 | 2.2 | 12.9 | 0.33 | -0.31 |
| 2.2 | 1.3 | 2.6 | 18.0 | -0.06 | -0.17 |
| 3.2 | 2.0 | 1.0 | 4.0 | 0.38 | 0.32 |
| 3.5 | 3.4 | 2.7 | 14.3 | -0.19 | 0.02 |
| 2.3 | 2.6 | 2.6 | 19.1 | -0.17 | -0.12 |
| 3.9 | 3.2 | 3.4 | 22 | -0.44 | -0.18 |
| -0.2 | 1.3 | 3.0 | 24.9 | -0.19 | -0.42 |
| -3.2 | -2.6 | 0.3 | -6.5 | 1.14 | 0.11 |

## V. MODEL IMPLEMENTATION

After feature selection and feature engineering, this section elaborates steps of model implementation. Two different sequential configurations, the construction of multi-time step dataset, the split of training and test set, the setup and tuning of network parameters and the forecasting process are explained as follows.

### A. Sequential Configuration

As a RNN neural network, LSTM and GRU each have three different sequential configurations: one-to-many, many-to-many and many-to-one [21, 25-26]. Many-to-many and many-to-one configurations are both suitable for sequence prediction problems. Their schematics are shown in Fig.6.

The proposed method aims to use three consecutive years' features to forecast the third year's future peak demand. Many-to-one configuration precisely captures this 3-to-1 relationship; in comparison, many-to-many configuration also includes two previous years' actual and forecasted peak demands. During the forecast process, since the actual peak demands of year 1 and year 2 are already known, producing forecasted values of year 1 and year 2 is not meaningful. However, the inclusion of the values of these two years does make a difference on the network loss function calculation during the network training process.

In order to match with the Mean Absolute Percentage Error (MAPE) later used for model evaluation in Section VI, Mean Absolute Error (MAE) is selected to construct the network loss functions. The loss function for many-to-one configuration is:

$$l_1 = \frac{1}{n}\sum_{j=1}^{n} \left| A^j - F^j \right| \tag{23}$$

where $n$ is the training batch size; $A^j$ is the third year's actual peak demand of $j_{th}$ record in the training batch; $F^j$ is the third year's forecasted peak demand of $j_{th}$ record in the training batch.

The loss function for many-to-many configuration is:

$$l_2 = \frac{1}{3n}\sum_{j=1}^{n}\sum_{i=1}^{3} \left| A_i^j - F_i^j \right| \tag{24}$$

where $n$ is the training batch size; $A_i^j$ is the $i_{th}$ year's actual peak demand of $j_{th}$ record in the training batch; $F_i^j$ is the $i_{th}$ year's forecasted peak demand of $j_{th}$ record in the training batch.

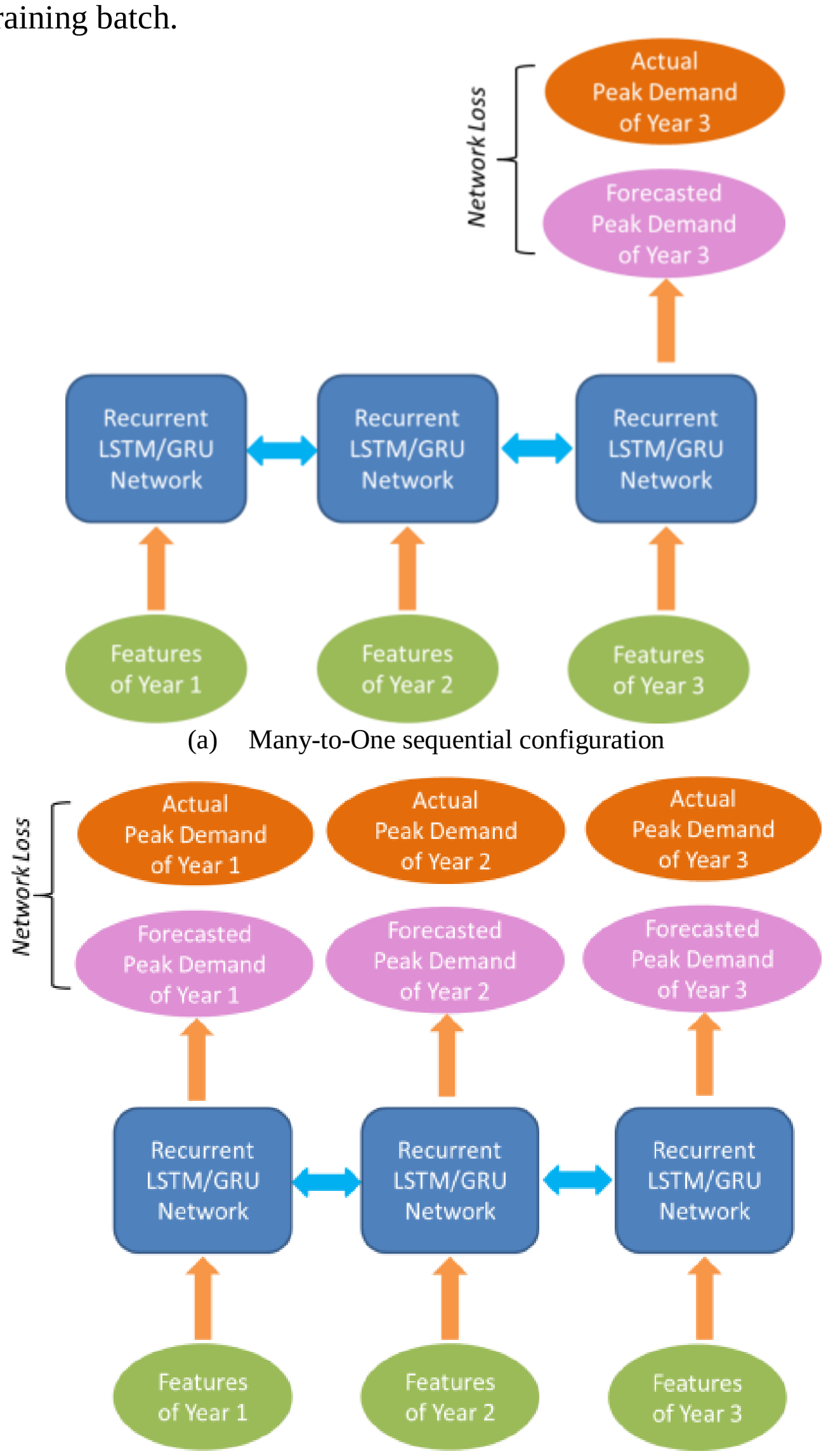

(a) Many-to-One sequential configuration

(b) Many-to-Many sequential configuration

Fig.6. Many-to-One and Many-to-Many sequential configuration

During the training process, each training batch's network loss is calculated and back propagated to update the weights between neurons. Theoretically, the advantage of many-to-one configuration is that its loss function $l_1$ is specific to the desired forecast target and the training should therefore make its best effort to minimize the error between the third year's actual and forecasted peak demand. However, when the actual peak demand fluctuates significantly within the three-year period, emphasizing only the third year's result may lead to a biased model. In comparison, many-to-many configuration's loss function $l_2$ measures the average of all three years' errors. When the output fluctuation is large, this configuration will be able to filter uncertainties and produce more consistent output year over year. The effect of these configurations will be further tested and discussed in Section VI.

### B. Multi-time Step Dataset

Different from traditional datasets used for FNN or other supervised learning models, LSTM and GRU models require data records to be grouped by a fixed number of time steps. This type of grouping is done for all feeders and all available years. Two dataset examples structured to many-to-one and many-to-many configurations are shown in Table III and Table IV.

Taking data record ID 1 in Table III and Table IV as an example, 2009/2010/2011 are three forecast years and 2011 is the final forecast year (Year-3) whose peak demand is the forecast target. The record has three rows and each row has features from a previous year and a forecast year. The previous year features are from 2008/2009/2010; the forecast year features are from 2009/2010/2011. It should be noted that the forecast year features in 2011 are forecasted values because in real application year-3 is a future year and its actual features are unknown. Comparing Table III and Table IV, the difference between many-to-one and many-to-many datasets is the inclusion of actual peak demand in 2009 and 2010 in this data record. This difference aligns with the network loss function difference explained in Section V-A.

### C. Training/Test Set Split

The multi-time step dataset should be randomly split into a model training set and test set by record. The training set is used to train the model; the test set is used to evaluate the model accuracy. A typical split ratio is 80% for training and 20% for testing. Model evaluation details will be discussed in Section VI.

### D. Network Parameter Setup and Tuning

Like a typical FNN, a LSTM/GRU network contains a specific number of hidden layers, a specific number of neurons in each hidden layer and a specific type of activation function for neurons. There is no direct way to determine these parameters rather than trying different combinations until acceptable results are obtained through model evaluation.

Optimization methods such as grid search and random search can be considered to facilitate the process [31]. Grid search initializes a finite set of reasonable values as the search space. For example, the number of hidden layers $\in \{1,2,3\}$ and the number of neurons $\in \{10,15,20\}$ can yield in total 9 combinations. Grid search goes through all these 9 combinations and chooses the combination with the best performance. Random search enumerates parameter combinations randomly instead of exhaustively traversing all of them. Other tuning algorithm such as Bayesian optimization is not recommended due to the limited size of our network.

### E. Forecasting Process

Once the model is established after training and evaluation, it can be used to forecast power demand in future years. Next year's forecasted economic and population features can be obtained from government or third-party agencies and are usually produced by their economists. It should be noted these forecast results could also contain errors. However, like in any multivariate LTLF methods, these numbers are often accurate enough, especially for the forecast of immediate coming years.

However, one feature that can hardly be forecasted accurately is the Max/Min Temperature of future years. To avoid this problem, planning engineers usually normalize future temperatures to a conservative value and produce future forecast using this common basis. For example, if the maximum temperature in the past decade is 35℃. When forecasting the next few years' loading, a safety margin 1℃ can be added to the historical high and 36℃ can be consistently used for forecast moving forward. Another benefit of the established forecasting model is that it can be used to retroactively normalize historical feeder peak demands using a statistical temperature so that historical yearly loadings can now be compared on the same basis. As a result, a more objective trend can be obtained.

Rare events such as World Cup games may lead to abnormal loading that cannot be effectively forecasted based on historical data. Extra safety margins can be added to the forecasted values in these cases.

The forecast can be performed continuously one year after another. For example, if 2019 is the first year to forecast, 2019's feeder peak demand will be firstly forecasted and then it is combined with 2018 and 2017 to forecast 2020's peak demand. This process continues until all years in a distribution planning horizon (e.g. next three years) are forecasted. It should be noted the accuracy level may gradually decrease but it is usually minor for the immediate next few years.

## VI. Application Example

The proposed approach was applied to a large urban grid in West Canada to establish both summer and winter long-term peak demand forecasting models for its distribution feeders that serve various types of loads.

### A. Description of Dataset

In total 289 distribution feeders were selected and their past 14-year annual data (2004-2017) were used to create the dataset. During this period, 182 of them had two-feeder or multi-feeder load transfer events among them and were converted to 87 virtual feeders. The remaining 107 feeders are actual feeders that either have no transfer events or have very short transfer periods that do not affect the correct gathering of annual peak demand. In total 1,997 valid three-year records were produced in the data format described in Table III and Table IV for both summer and winter. In order to reveal models' true forecasting capability, for every third year, instead of using the actual values, forecasted economic and population features prior to that year were used. The 1,997 records were split into 1,597 records for training and 400 records for testing based on the 80%/20% split ratio. To evaluate the model's forecast accuracy, the trained model was tested on the 400 test records and compared to the actual peak demand values.

TABLE III: DATASET EXAMPLE FOR SUMMER PEAK DEMAND FORECAST (MANY-TO-ONE CONFIGURATION)

| Data Record ID | Feeder ID | Forecast Year | Previous Year Features | | | Forecast Year Features | | | | | Year-3 Peak Demand (Year 3) |
|---|---|---|---|---|---|---|---|---|---|---|---|
| | | | Previous-Year Peak Demand | Residential Peak Load percentage | Commercial Peak Load percentage | EP1 | EP2 | Maximum Temperature | Maximum Temperature Change | Large Customer Net Load Change | |
| 1 | 1001 | 2009 | 433 A | 66.5% | 10.2% | -0.64 | 0.44 | 33.3℃ | 0.7℃ | 42 A | 550 A (2011) |
| | 1001 | 2010 | 502 A | 63.1% | 11.1% | -0.16 | 0.31 | 32.0℃ | -1.3℃ | 34 A | |
| | 1001 | 2011 | 554 A | 63.0% | 11.3% | 0.33 | -0.31 | 35.4℃ | 3.4℃ | 0 A | |
| 2 | 1001 | 2010 | 502 A | 63.1% | 11.1% | -0.16 | 0.31 | 32.0℃ | -1.3℃ | 34 A | 521 A (2012) |
| | 1001 | 2011 | 554 A | 63.0% | 11.3% | 0.33 | -0.31 | 35.4℃ | 3.4℃ | 0 A | |
| | 1001 | 2012 | 540 A | 59.4% | 12.7% | -0.06 | -0.17 | 33.2℃ | -2.2℃ | -21 A | |
| … | … | … | … | … | … | | | … | … | … | … |
| 238 | 1321 | 2010 | 317 A | 94.2% | 5.8% | -0.16 | 0.31 | 32.0℃ | -1.3℃ | 0 A | 331 A (2012) |
| | 1321 | 2011 | 323 A | 93.8% | 6.2% | 0.33 | -0.31 | 35.4℃ | 3.4℃ | 10 A | |
| | 1321 | 2012 | 329 A | 94.1% | 5.9% | -0.06 | -0.17 | 33.2℃ | -2.2℃ | 0 A | |
| | … | **…** | | … | … | … | … | … | … | … | … |

TABLE IV: DATASET EXAMPLE FOR SUMMER PEAK DEMAND FORECAST (MANY-TO-MANY CONFIGURATION)

| Data Record ID | Feeder ID | Forecast Year | Previous Year Features | | | Forecast Year Features | | | | | Forecast Year Peak Demand |
|---|---|---|---|---|---|---|---|---|---|---|---|
| | | | Previous-Year Peak Demand | Residential Peak Load percentage | Commercial Peak Load percentage | EP1 | EP2 | Maximum Temperature | Maximum Temperature Change | Large Customer Net Load Change | |
| 1 | 1001 | 2009 | 433 A | 66.5% | 10.2% | -0.64 | 0.44 | 33.3℃ | 0.7℃ | 42 A | 502 A |
| | 1001 | 2010 | 502 A | 63.1% | 11.1% | -0.16 | 0.31 | 32.0℃ | -1.3℃ | 34 A | 554 A |
| | 1001 | 2011 | 554 A | 63.0% | 11.3% | 0.33 | -0.31 | 35.4℃ | 3.4℃ | 0 A | 550 A |
| 2 | 1001 | 2010 | 502 A | 63.1% | 11.1% | -0.16 | 0.31 | 32.0℃ | -1.3℃ | 34 A | 554 A |
| | 1001 | 2011 | 554 A | 63.0% | 11.3% | 0.33 | -0.31 | 35.4℃ | 3.4℃ | 0 A | 550 A |
| | 1001 | 2012 | 550 A | 59.4% | 12.7% | -0.06 | -0.17 | 33.2℃ | -2.2℃ | -21 A | 521 A |
| … | … | … | … | … | … | | | … | … | … | … |
| 238 | 1321 | 2010 | 317 A | 94.2% | 5.8% | -0.16 | 0.31 | 32.0℃ | -1.3℃ | 0 A | 323 A |
| | 1321 | 2011 | 323 A | 93.8% | 6.2% | 0.33 | -0.31 | 35.4℃ | 3.4℃ | 10 A | 329 A |
| | 1321 | 2012 | 329 A | 94.1% | 5.9% | -0.06 | -0.17 | 33.2℃ | -2.2℃ | 0 A | 331 A |
| | … | **…** | | … | … | … | … | … | … | … | … |

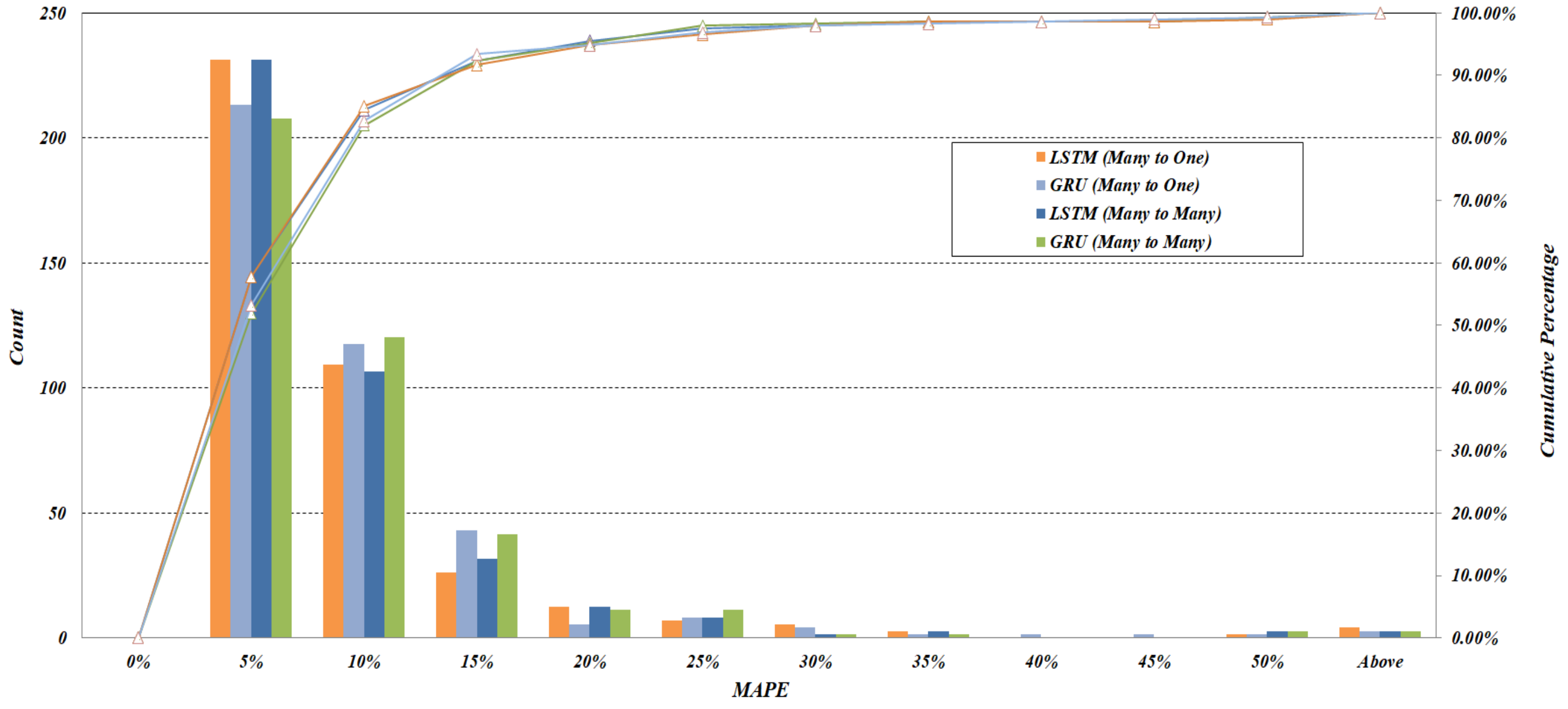

Fig.7. MAPE distribution in summer

TABLE V: SUMMER RESULTS

| | LSTM(Many-to-One) | GRU (Many-to-One) | LSTM(Many-to-Many) | GRU(Many-to-Many) |
|---|---|---|---|---|
| MAPE(%) | 6.67 | 6.92 | 6.54 | 6.79 |
| Cumulative Percentage with MAPE≤10% (%) | 85.12 | 82.70 | 84.43 | 82.01 |
| Model Training Time (s, Epoch=200) | 202.46 | 164.09 | 200.33 | 162.28 |

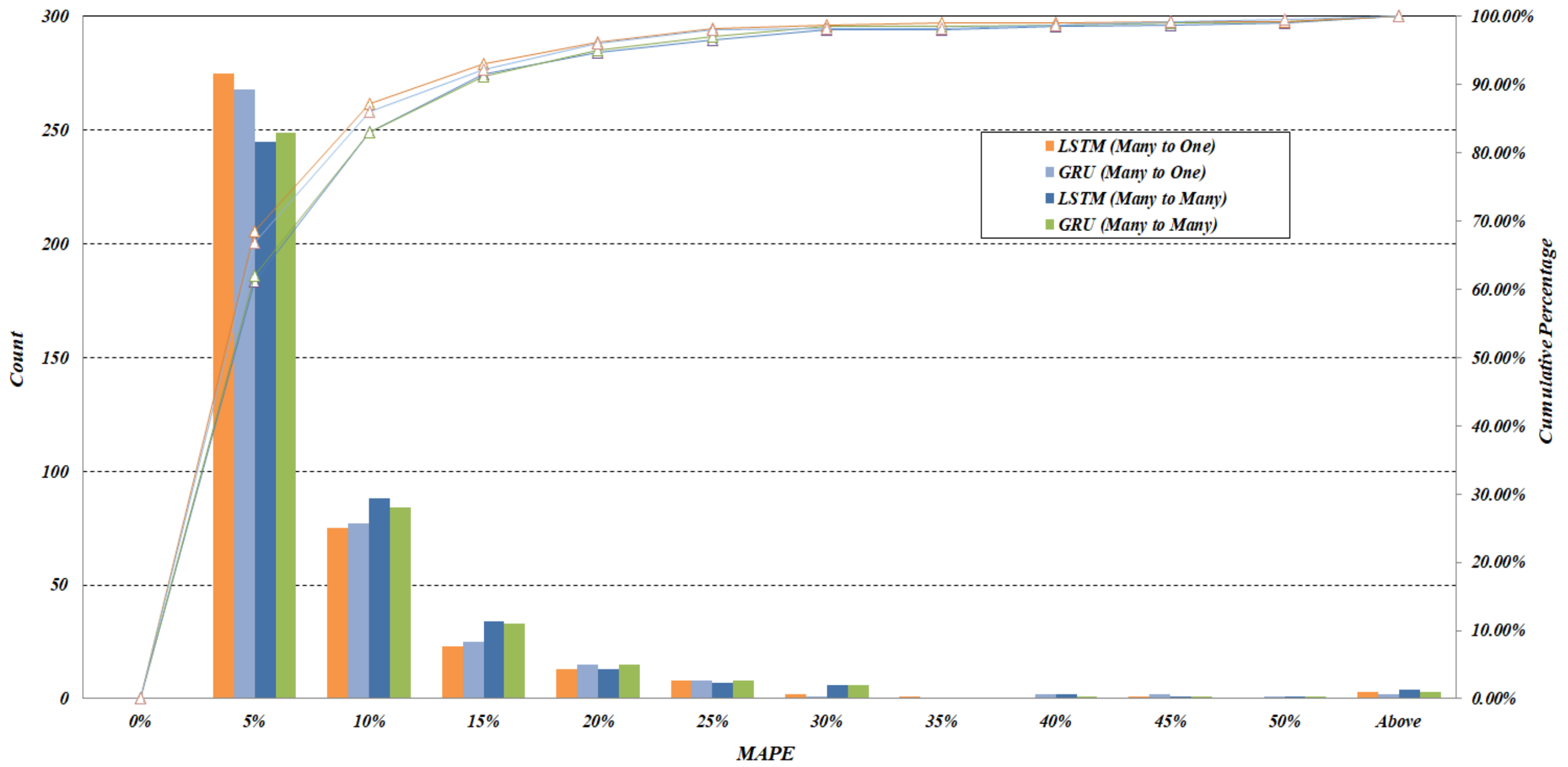

Fig.8. MAPE distribution in winter

TABLE VI: WINTER RESULTS

| | LSTM(Many-to-One) | GRU (Many-to-One) | LSTM(Many-to-Many) | GRU(Many-to-Many) |
|---|---|---|---|---|
| MAPE(%) | 5.09 | 5.20 | 6.15 | 6.08 |
| Cumulative Percentage with MAPE≤10% (%) | 87.28 | 86.03 | 83.04 | 83.04 |
| Model Training Time (s, Epoch=200) | 200.11 | 161.20 | 203.85 | 163.59 |

*B. Results of LSTM and GRU under Two Sequential Configurations*

LSTM and GRU models were implemented with both many-to-many and many-to-one configurations for summer and winter separately. Each configuration uses 8 input features as listed in Table III and IV. Summer results are summarized in Fig.7 and Table V; winter results are summarized in Fig.8 and Table VI. For each model, two hidden layers were composed. One hidden layer is a LSTM/GRU layer which contains hidden states and the other hidden layer is a regular neural network layer connecting to the output layer. Each hidden layer contains 6 neurons. The rectified linear unit (ReLU) activation function is used in each neuron. MAPE was chosen as the first accuracy metric. It is commonly used for measuring prediction error and is given as:

$$MAPE = \frac{1}{m}\sum_{i=1}^{m}\left|\frac{A_i - F_i}{A_i}\right| \times 100\% \qquad (25)$$

where $m$ is the total number of test records; $F_i$ is the forecasted value of $i_{th}$ record; $A_i$ is the actual value of $i_{th}$ record.

Based on the MAPE results, histograms are produced to present the error distributions in Fig.7 and Fig.8. The records with less than or equal to 10% MAPEs are counted and its percentage number against the total number of records (i.e. 400) is calculated. This percentage number directly reflects how many forecasted records have high confidence levels.

Another aspect of the evaluation is in regards to the model training time. Once a model is trained, the time required for testing using the trained model is almost negligible and is therefore not included for comparison. The hardware computational environment for this application example is Intel i7-6700K CPU @ 4.00GHz with 4 Cores and 16GB DDR4 RAM memory. The software environment is Windows 10 and Python 3.5 with Tensorflow backend. The total number of training epochs was set to 200 with a batch size of 10.

A few observations can be drawn from the obtained results:

- Overall, all models and configurations are quite accurate. This shows the great value of the proposed method. It was found that most large errors are attributed to abnormal load behaviors during two dramatic economic downturns in 2009 and 2015-2016 in the interested region. Prior to these downturns, no one forecasted the economy and population features correctly and the input errors lead to the power demand forecast errors in the results.
- LSTM and GRU have very close accuracy levels. From MAPE and Cumulative Percentage metrics, LSTM slightly outperformed GRU in summer in both configurations while GRU slightly outperformed LSTM in winter under the many-to-many configuration. The differences between them are very small.
- In winter, many-to-one configurations noticeably outperformed many-to-many configurations with both

LSTM and GRU models. However, in summer the accuracy difference becomes marginal and many-to-many configuration becomes slightly better than many-to-one configuration in MAPE (%). One explanation for this interesting observation is that the summer loading records in the historical timespan have larger fluctuations due to the rapidly increasing use of air conditioners year over year in the region. On the contrary, natural gas is the main source of heating in winter in the region and does not consume much electricity. Therefore, if economic and population changes are excluded, the loading of winter months between adjacent years is relatively stable. In Section V, it has been discussed that many-to-one configuration only focuses on the desired output and can be more accurate if the yearly fluctuation is low. This is the reason many-to-one configuration is noticeably more accurate in winter. In comparison, many-to-many configurations are more immune to output fluctuation because its network loss function automatically smooth out the differences within three consecutive years and therefore produces a more neutral forecast model through training. This is the reason the advantage of many-to-many configuration is revealed in summer while the advantage of many-to-one configuration is undermined.

- From the training time perspective, GRU is significantly faster than LSTM. This was expected because of fewer gates used in the model. The time difference between many-to-many and many-to-one configurations is negligible. This is because the computation time required for calculating different loss functions is not much different in the given computational environment. Since LTLF is not performed in real-time and all the listed training times are generally acceptable, it is recommended that all 8 models are tested and the best models get selected for future summer and winter forecast separately.

### *C. Improvement By Using Virtual Feeders*

The most accurate model and configuration by MAPE (%) from Table V and VI were selected and tested without the use of virtual feeders. Their MAPE results are shown in Table VII.

TABLE VII: IMPROVEMENT BY USING VIRTUAL FEEDERS

| Use Virtual Feeder Features? | Summer: LSTM Many-to-Many MAPE (%) | Winter: LSTM Many-to-One MAPE (%) |
|---|---|---|
| No | 14.27% | 11.48% |
| Yes | 6.54% | 5.09% |

Significant performance improvement can be observed after using the proposed virtual feeder features to eliminate data noise caused by the load transfer events.

### *D. Comparison to Other Models*

As part of the model evaluation, the proposed model was compared to various traditional models established as below. Virtual feeder features are also used in these models.

- Bottom-up model: As discussed in Section I, only Large Customer Net Load Change feature was gathered and added to the Previous-Year Peak Demand to calculate the following year's peak demand.
- ARIMA model: The same 14-year loading data of 289 feeders were used. For each feeder, its first 13-year peak demand values were fed into a ARIMA (2,0,0) model for training. ARIMA (2,0,0) was found to give the best forecast result among different ARIMA order parameters for this dataset. Then the peak demand values in 2017 were forecasted and compared to the true values to calcualte MAPE.
- One-year FNN: The same 14-year data of 289 feeders were used. For each feeder, only one year features are used to forecast. Each training record is like each row in Table IV. A traditional FNN model is used. The input layer has 8 neurons (features) and two hidden layers each have 6 neurons. ReLU activation functions are used in the hidden and output layers.
- Three-year FNN: The same 14-year data of 289 feeders were used. Instead of using the LSTM/GRU neural network, a traditional FNN model is used to incorporate all the features of three consecutive years (in total 24 features) to forecast the third year's peak demand. The input layer has 24 neurons and two hidden layers each have 12 neurons. ReLU activation functions are used in the hidden and output layers.

TABLE VIII: PERFORMANCE COMPARISON OF DIFFERENT MODELS

| Model | Summer MAPE (%) | Winter MAPE (%) |
|---|---|---|
| Best LSTM/GRU Models | 6.54% | 5.09% |
| Bottom-up | 16.61% | 14.80% |
| ARIMA | 14.51% | 11.33% |
| One-year FNN | 14.62% | 10.19% |
| Three-year FNN | 8.51% | 7.88% |

As shown in Table VIII, the proposed model outperformed all other models for both summer and winter forecasting.

## VII. CONCLUSIONS

This paper presents a novel and comprehensive method for distribution feeder long-term load forecast. Compared to previous methods, the advantages of this method are:

- It uses a hybrid model which can seamlessly integrate different levels of information including both top-down and bottom-up features. It can therefore capture the relationships among feeder peak demand, overall regional drivers and individual feeder load details.
- It uses advanced sequence prediction models LSTM and GRU to effectively capture and leverage the sequential characteristics of multi-year data to improve forecast accuracy.

The proposed method was applied to a large urban grid in West Canada. LSTM and GRU models under two sequential configurations and a few different traditional models were all implemented and compared in detail. The proposed method with the use of virtual feeder demonstrated superior performance for both summer and winter forecasts compared to traditional models. Overall, LSTM and GRU models have similar performances. GRU model is faster than LSTM model. The performances of many-to-many and many-to-one sequential configurations are affected by the output fluctuation level between adjacent years. Since the training times are not very long, it is recommended to evaluate both LSTM and GRU models under two sequential configurations and choose the best performers for forecast application.

**Ming Dong** (S’08, M’13, SM’18) received his Ph.D degree from Department of Electrical and Computer Engineering, University of Alberta, Canada in 2013. Since graduation, he has been working in various roles in two major electric utility companies in West Canada as a registered Professional Engineer (P.Eng.) and Senior Engineer for 6 years. Dr. Dong was the recipient of the Certificate of Data Science and Big Data Analytics from Massachusetts Institute of Technology. He is also a regional officer of Alberta Artificial Intelligence Association. His research interests include applications of artificial intelligence and big data technologies in power system planning and operation, power quality data analytics and non-intrusive load monitoring.

**Lukas Grumbach** is the co-founder and Chief Data Scientist of Auroki Analytics, a data consulting company located in Vancouver, Canada. Lukas holds a Bachelor of Science degree in Mathematics from University of Basel and a Master of Science degree in Computer Science from Swiss Federal Institute of Technology Lausane, Switzerland. He has 10 years of mathematical modeling and data science experience for various clients in the oil and gas, finance and public utility industries.